\documentclass[aps,physrev,preprint,superscriptaddress]{revtex4-2}
\usepackage{graphicx}
\usepackage{dcolumn}
\usepackage{bm}
\usepackage{epsfig,graphics,amssymb,amsmath,subeqnarray,mathrsfs,color,xcolor, rotating, mhchem}
\usepackage{siunitx}
\usepackage{float}
\makeatletter
\let\newfloat\newfloat@ltx
\makeatother
\usepackage{algorithm}
\usepackage{algpseudocode}
\usepackage{algorithmicx}
\usepackage{amssymb}
\usepackage{dutchcal}

\def\e{\mathbf{e}}
\def\u{\mathbf{u}}
\def\x{\mathbf{x}}
\def\t{\mathbf{t}}
\def\n{\mathbf{n}}
\def\X{\mathbf{X}}
\def\f{\mathbf{f}}
\def\F{\mathbf{F}}
\def\M{\mathbf{M}}

\def\p{\mathbf{p}}

\begin{document}

\preprint{}

\title{Navigation of a Three-Link Microswimmer via Deep Reinforcement Learning}


\author{Yuyang Lai}
\affiliation{Department of Mechanics and Engineering Science at College of Engineering, Beijing 100871, PR China}

\author{Sina Heydari}
\affiliation{Department of Mechanical Engineering, Santa Clara University, Santa Clara, CA 95053, USA}

\author{On Shun Pak}
\email[Email address for correspondence:]{opak@scu.edu}
\affiliation{Department of Mechanical Engineering, Santa Clara University, Santa Clara, CA 95053, USA}
\affiliation{Department of Applied Mathematics, Santa Clara University, Santa Clara, CA 95053, USA}

\author{Yi Man}
\email[Email address for correspondence:]{yiman@pku.edu.cn}
\affiliation{Department of Mechanics and Engineering Science at College of Engineering, Beijing 100871, PR China}




\date{\today}

\begin{abstract}
Motile microorganisms develop effective swimming gaits to adapt to complex biological environments. Translating this adaptability to smart microrobots presents significant challenges in motion planning and stroke design. In this work, we explore the use of reinforcement learning (RL) to develop stroke patterns for targeted navigation in a three-link swimmer model at low Reynolds numbers. Specifically, we design two RL-based strategies: one focusing on maximizing velocity (Velocity-Focused Strategy) and another balancing velocity with energy consumption (Energy-Aware Strategy). Our results demonstrate how the use of different reward functions influences the resulting stroke patterns developed via RL, which are compared with those obtained from traditional optimization methods. Furthermore, we showcase the capability of the RL-powered swimmer in adapting its stroke patterns in performing different navigation tasks, including tracing complex trajectories and pursuing moving targets. Taken together, this work highlights the potential of reinforcement learning as a versatile tool for designing efficient and adaptive microswimmers capable of sophisticated maneuvers in complex environments.
\end{abstract}


\maketitle

\section{Introduction}
Locomotion at low-Reynolds numbers is a fascinating subject, as the interaction between microorganisms and their environment generates propulsion in ways fundamentally different from macroscopic motion \cite{lauga2009hydrodynamics,elgeti2015physics,yeomans2014introduction}. Microorganisms navigate their viscous environments through specialized mechanisms, such as the undulating flagella of sperm cells \cite{fauci2006biofluidmechanics}, the rotating helical flagella of bacteria \cite{lauga2016bacterial}, and the coordinated ciliary movements of paramecia \cite{parducz1967ciliary}. Inspired by these natural strategies, various microswimmers have been designed for applications such as drug delivery \cite{gao2012cargo,ceylan20193d,zhang2009characterizing}, self-assembly \cite{grosjean2015remote,cheang2015self}, and targeted therapy \cite{huang20153d}. A core challenge in the design of microswimmers is the development of effective stroke patterns or motion planning: what body deformations can achieve the desired locomotion? Unlike microorganisms, which can adapt their gaits based on environmental cues and functional needs, most current microswimmers possess a single mode of motion and can only operate in simple, controlled environments \cite{gao2012cargo,hu2018small,ohm2010liquid,dai2016programmable,palagi2020soft,von2018gait}. Addressing this challenge requires not only an understanding of the biomechanics of microbial movement but also insights into how their detailed structures and sensory systems coordinate to achieve their goals, making the modeling process inherently complex \cite{nassif2002extracellular,celli2009helicobacter,mirbagheri2016helicobacter}.

Model-free reinforcement learning (RL) offers a promising approach for stroke design and motion planning in microswimmers. Recent computational and experimental studies have demonstrated the potential of RL in studying biophysical problems at low-Reynolds numbers and designing intelligent microswimmers \cite{tsang2020self,jiao2021learning,qiu2022navigation,zou2022gait,zhu2022optimizing,qin2023reinforcement}. Within the RL framework, microswimmers learn from experience through trial and error without relying on physical knowledge of the system. This allows for the discovery of novel locomotion strategies that traditional modeling approaches may not easily uncover. For example, RL has enabled microswimmers to achieve targeted navigation, adapting their movements in response to complex environmental cues and disturbances, ensuring robust performance even in dynamic and unpredictable fluid environments \cite{zou2022gait,colabrese2017flow,alageshan2020machine,schneider2019optimal,muinos2021reinforcement,yang2020micro,yang2024machine}. Studies have shown that microswimmers can optimize their swimming strategies to achieve specific goals, such as maximizing speed or efficiency, by adjusting their stroke patterns accordingly \cite{zou2022gait,zhu2022optimizing,qin2023reinforcement}. Additionally, RL has been successfully applied to scenarios involving multiple microswimmers, facilitating coordinated behaviors such as pursuit-evasion dynamics and collective navigation, which are critical for applications like targeted drug delivery and environmental sensing \cite{zhu2022optimizing,liu2023learning}. These advancements demonstrate how reinforcement learning can effectively address the challenges associated with microswimmer design, offering a powerful tool for developing efficient and intelligent micro-robots capable of performing sophisticated tasks in complex biological environments. \cite{muinos2021reinforcement,amoudruz2022independent,behrens2022smart}.

In this work, we consider a three-link swimmer, one of the simplest microswimmer models capable of generating propulsion at low-Reynolds numbers. We utilize RL to explore the development of stroke patterns for targeted navigation. We design two strategies—one focusing on maximizing velocity (Velocity-Focused Strategy) and another balancing velocity with energy consumption (Energy-Aware Strategy). We examine the stroke patterns developed through RL based on different reward functions. Our results underscore the effectiveness and versatility of RL in developing stroke patterns to meet various performance goals, demonstrating the potential for RL as a tool to design locomotory gaits of microswimmers. We also showcase the capability of the RL-powered microswimmer in performing complex navigation tasks in scenarios relevant to their potential biomedical applications.

This paper is structured as follows. In \S \ref{sec:model}, we introduce the three-link swimmer model, detailing its degrees of freedom and its dynamics at low-Reynolds number. \S \ref{sec:RL} describes the RL framework we employed, including the design of the two strategies: the Velocity-Focused Strategy and the Energy-Aware Strategy. We outline the neural network architecture and the reward functions tailored for each strategy. In \S \ref{sec:results}, we present the results of our simulations, analyzing the swimmer's performance under both strategies. We compare the stroke patterns developed through RL with those from previous studies, highlighting similarities and differences.  Additionally, we demonstrate the RL framework's capability to develop complex stroke patterns for tracing a star-shaped trajectory and navigating toward moving targets. We conclude this work with remarks on its limitations in \S \ref{sec:conclusion}.
\section{Model of a three-link swimmer} \label{sec:model}

\begin{figure*}[t]
\centering
 	\includegraphics[scale=0.8]{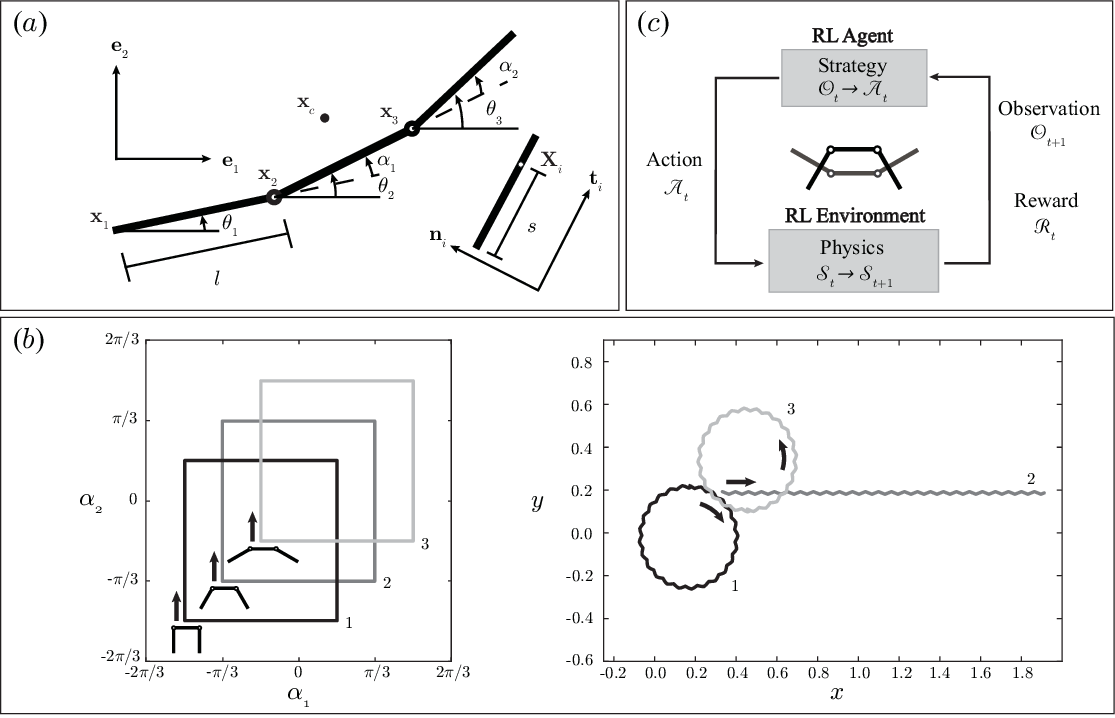}
 	\caption{\footnotesize $(a)$ Model of three-link swimmer. It consists of three rigid links of equal length, which are connected by two hinges, allowing rotation to adjust the relative angles $\alpha_i$ $(i = 1, 2)$. The swimmer's geometric centroid denoted $\mathbf{x}_c$, serves as the reference point for its motion. $(b)$ Three basic stroke patterns of the three-link swimmer. Left: stroke patterns in phase space; Right: corresponding trajectories of the geometric centroid in physical space. The initial configurations for these movements are shown in the left panel. $(c)$ Framework of model-free reinforcement learning. } 
 	\label{fig:3linkswimmer_RL}
\end{figure*} 
The three-link swimmer possesses the minimal degrees of freedom required for self-propulsion in a low-Reynolds-number environment \cite{purcell1977life}. This system consists of three identical rigid links, each with a radius $a$ and a length $l = L/3$, where $L$ represents the total length of the swimmer (see Fig.~\ref{fig:3linkswimmer_RL}$a$). The locomotion of the swimmer is constrained to two dimensions, described using the Cartesian coordinates $(\e_1,\e_2)$. The left end of each link is denoted by $\x_i = (x_i, y_i)$, and its orientation by $\t_i$. The angle between $\t_i$ and $\e_1$ is represented by $\theta_i$. The swimmer's hinges allow for free rotation, with the angles between adjacent links denoted as $\alpha_1$ and $\alpha_2$. By actuating these angles, the links interact with the surrounding fluid, resulting in net propulsion. To avoid close proximity, the angles $\alpha_1$ and $\alpha_2$ are restricted to the range $[-2\pi/3, 2\pi/3]$.


The position of any point on link $i$ is denoted by $\X_i = \x_i + s\t_i$, where $s$ represents the distance along the link from its left end. The local velocity at $\X_i$ is given by:
\begin{align}\label{eq:localvelocity}
\u_i = \dot\x_i+s\dot\theta_i \n_i, 
\end{align}
where $\n_i$ represents the unit vector normal to link $i$. Based on the resistive force theory, the local hydrodynamic force is proportional to the local velocity \cite{lighthill1975mathematical}. Consequently, the local force is calculated as follows:
\begin{align}\label{eq:localforce}
    \f_i=-\left(C_{\parallel}\t_i\t_i+C_\perp\n_i\n_i\right)\cdot\u_i,
\end{align}
where $C_\parallel=2\pi \mu/\left[\ln\left(L/a\right)-1/2\right]$ and $C_\perp=4\pi \mu/\left[\ln\left(L/a\right)+1/2\right]$ are the drag coefficients \cite{lighthill1975mathematical}, and $\mu$ is the dynamic viscosity of the fluid. Integrating along link $i$, the total force and hydrodynamic torque are:
\begin{align}
    \F_i=\int_0^l\f_i\:\mathrm{d}s, \quad \M_i&=\int_0^l\X_i\times\f_i\mathrm{d}s.
\end{align}

For low-Reynolds-number locomotion, the total hydrodynamic force and torque on the swimmer should vanish, namely
\begin{align}\label{eq:forcebalance}
\sum_{i=1}^{3}\F_{i}=\mathbf{0}, \quad \sum_{i=1}^{3}\M_i=\mathbf{0}.
\end{align}
Moreover, the motion of the swimmer has kinematic constraints (here $i=1,2$):
\begin{align}
\label{eq:constraints}
\x_{i+1}-\x_{i}=l\t_i, \quad \theta_{i+1}-\theta_i=\alpha_i.
\end{align}
In presenting our results, we scale all lengths by the total length of the swimmer, $L$. We assume a characteristic time scale, $T_0$, which corresponds to the actuation rate of the angle between neighboring links. The associated force scale is defined as $C_\perp L^2/T_0$. As a result, the dimensionless quantities are defined as $\x_i=L\overline\x_i,\ \overline{\dot{\alpha_j}} =T_0 \dot\alpha_j,\ \gamma = C_\parallel/C_\perp$, where $i = 1,2,3$ and $j = 1,2$. In this study we consider a slender swimmer ($a\ll L$) with $\gamma = 1/2$. To simplify the notations, we omit the overbars hereafter and refer only to dimensionless quantities. Combining Eqs.~\eqref{eq:forcebalance} and \eqref{eq:constraints}, the swimmer's motion is described by a system of linear equations:
\begin{align}\label{eq:motion}
    \boldsymbol{H}(\boldsymbol{X},\boldsymbol{Y},\boldsymbol{\Theta})
\begin{pmatrix}
  \dot{\boldsymbol{X}} \\
  \dot{\boldsymbol{Y}} \\
  \dot{\boldsymbol{\Theta}}
\end{pmatrix}=\boldsymbol{q},
\end{align}
where $\boldsymbol{X}=[x_1,x_2,x_3]^\top,\boldsymbol{Y}=[y_1,y_2,y_3]^\top,\boldsymbol{\Theta}=[\theta_1,\theta_2,\theta_3]^\top$, while $(\dot{\boldsymbol{X}},\dot{\boldsymbol{Y}},\dot{\boldsymbol{\Theta}})$ are their derivative with respect to time $t$. The vector $\boldsymbol{q}$ is the function of actuation rates of the angle between neighboring links $\dot{\alpha}_1,\dot{\alpha}_2$. See Supplementary Materials~\cite{supplement} for the components of $\boldsymbol{H}$ and $\boldsymbol{q}$.

All instantaneous configurations of the swimmer can be represented by a point in the two-dimensional $(\alpha_1,\alpha_2)$ phase space. Thus, all periodic stroke patterns of the swimmer can be depicted as a single closed curve in this space. In Fig.~\ref{fig:3linkswimmer_RL}$(b)$, we illustrate three stroke patterns in the phase space (left panel) along with the corresponding trajectories of the swimmer's geometric centroid in the physical space (right panel). The classical Purcell's stroke pattern is shown in gray lines. In this pattern, only one arm moves at a time, maintaining symmetry with joint angles ranging from $-\pi/3$ to $\pi/3$. This symmetric stroke results in the swimmer moving straight along the horizontal direction. We modify Purcell's stroke pattern by allowing the joint angles to vary asymmetrically between $-\pi/2$ and $\pi/6$, as illustrated by the light gray lines. This asymmetry causes the swimmer to move along a clockwise circular trajectory. Similarly, if the joint angles vary between $-\pi/6$ and $\pi/2$, shown by the black lines, the swimmer moves along a counterclockwise circular path.

\section{Targeted navigation via reinforcement learning}\label{sec:RL}

\subsection{RL framework for targeted navigation}

We use a reinforcement learning (RL) framework to train the swimmer in swimming parallel along a certain target direction $\theta_T$ (Fig.~\ref{fig:3linkswimmer_RL}c). The state of the system, $\mathcal{S}\in (\x_1,\theta_1,\theta_2,\theta_3)$, is specified by the coordinate of swimmer's one end $\boldsymbol{x}_1$ and links orientations $\theta_1,\theta_2,\theta_3$. The observation,  $\mathcal{O}\in(\cos\theta_d,\sin\theta_d,\alpha_1,\alpha_2)$, is extracted from the state, where $\theta_d=\theta_2-\theta_T$ is the difference between the second link's orientation and the target direction. The term $(\cos\theta_d,\sin\theta_d)$ is introduced to ensure continuity in the orientation space, as each component remains within $\left[-1,1\right]$. This ensures that our data will not overflow, thereby preventing the continuity of the values from being disrupted when taking $\theta_d$ modulo $2\pi$. The agent in the RL framework utilizes an Actor-Critic neural network architecture to decide the swimmer's actions based on the observations. Specifically, for each action step, the swimmer senses its observation $\mathcal{O}$ and, through the Actor network, determines the action $\mathcal{A} \in (\dot{\alpha}_1, \dot{\alpha}_2)$ by calculating the angular velocities for rotating its two hinges.

We design different reward functions to evaluate the success of the swimmer's actions in achieving targeted navigation. Two types of objective criteria are established for control. The first objective focuses on velocity toward the target direction, which we refer to as the Velocity-Focused Strategy (VFS). The criterion here is the distance traveled by the swimmer along the target direction within a specific time period. Specifically, the reward function for VFS is defined as follows:
\begin{align}
    \mathcal{R}_k=b(\x_{c_{k+1}}-\x_{c_k})\cdot\p,
\end{align}
where $k$ represents the ordinal number of training step, and $\x_{c_k}$ denotes the geometric centroid of the swimmer at the $k$th training step. The targeted orientation is denoted as $\p = \cos{\theta_T} \e_1 + \sin{\theta_T} \e_2$. 
The parameter $b$ is a positive scaling factor introduced to adjust the magnitude of the reward signal. A larger value of $b$ increases the reward's magnitude, which can accelerate the convergence rate of training by encouraging larger updates during gradient descent. However, if $b$ is set too high, it may lead to numerical instability due to excessively large gradient steps. Therefore, $b$ should be chosen carefully to balance the trade-off between faster convergence and stable learning dynamics.

The second objective is to achieve an Energy-Aware Strategy (EAS), which aims to realize targeted navigation while penalizing energy consumption.  We consider the total rate of work done by the swimmer on the fluid:
\begin{align}\label{eq:energy}
    \Phi=\sum_{i=1}^3\Phi_i,
\end{align}
where $\Phi_i$ refers to the rate of work done by the $i$th link and can be computed as follows:
\begin{align}
    \Phi_i&=\int_{0}^{1/3}-\f_i\cdot \u_i\:\mathrm{d}s,
\end{align}
where $\f_i$ and $\u_i$ are given by Eqs.~\eqref{eq:localvelocity} and \eqref{eq:localforce}.

In the actual training process, we calculate the work done by the swimmer during the $k$th training step, defining it as:
\begin{align}
W_k = \int_{t_k}^{t_{k+1}} \Phi \:\mathrm{d}t,
\end{align}
where $\t_k$ denotes the initial time of the $k$th training step. By incorporating an energy penalty, we design the reward function for the EAS as:
\begin{align}
 \mathcal{R}_k=b(\x_{c_{k+1}}-\x_{c_k})\cdot\p-c W_k,
\end{align}
where $c$ is a positive weight introduced to penalize mechanical power consumption during each action step. A larger $c$ increases the emphasis on reducing energy expenditure, which can lead to higher swimming efficiency. However, if $c$ is set too high, the swimmer may prioritize conserving energy over progressing towards the target, resulting in decreased accuracy in navigating along the desired direction or even failure to reach the target.

\subsection{Training process}

\begin{figure*}[t]
\centering
 	\includegraphics[scale=0.8]{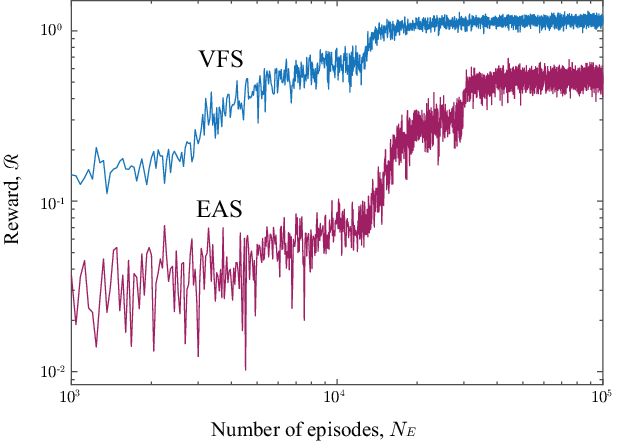}
 	\caption{\footnotesize Convergence of reward functions for the Velocity-Focused Strategy (VFS, blue line) and the Energy-Aware Strategy (EAS, purple line). Each training episode contains a fixed number of action steps $N_s=200$. The reward $\mathcal{R} = \sum_{k=1}^{N_s} \mathcal{R}_k$ denotes the cumulative reward obtained over all action steps within a single episode.}
 	\label{fig:rewardvsepisode}
\end{figure*}

We employ the Proximal Policy Optimization (PPO) algorithm to train the swimmer to navigate along a specified target direction, $\theta_T$. The algorithm is adapted from \cite{jiao2021learning, zou2022gait} (see Supplementary Materials~\cite{supplement} for more details). Without loss of generality, we set the target direction to be parallel to the $x$-direction, corresponding to a target angle of $\theta_T=0$.
To fully explore the observation space $\mathcal{O}$, $\left(\theta_1,\theta_2,\theta_3\right)$ in the initial swimmer state $\mathcal{S}$ are randomized at the beginning of each episode. 
The training process is divided into $N_E$ episodes, each consisting of $N_s$ action steps. A suﬀiciently large number of episodes and action steps is necessary to ensure the convergence of the training results and smoothness of the swimmer’s movements.  In the reward functions, we set the coefficients to $b = 6$ and $c = 3$. Our numerical experiments show that choosing a value of  $b <6$ increases the convergence time, though the training results remain similar to when $b=6$. However, increasing $b$ beyond 6 may cause numerical instability due to larger gradient steps,  resulting in deviations from the target direction during navigation. The coefficient $c$ is a positive weight that penalizes mechanical power consumption, which may be expected to reduce performance in terms of displacement toward the target direction or the ability to reorient toward it. When $c$ exceeds 3, we observe a significant asymmetry in the stroke patterns, rendering the navigation strategy ineffective. This occurs because the swimmer prioritizes energy conservation over advancing toward the target, leading to a decrease in navigation accuracy (refer to the Supplemental Material~\cite{supplement} for more details on the effects of these parameters).




In Fig.~\ref{fig:rewardvsepisode}, we compare the progression of rewards versus the number of training episodes for both the VFS and the EAS reward functions. Here, the reward $\mathcal{R} = \sum_{k=1}^{N_s} \mathcal{R}_k$ denotes the cumulative reward obtained over all action steps within a single episode. It can be observed that while both training processes eventually converge, the EAS requires more episodes to do so. Specifically, the VFS rewards converge around 15,000 episodes, whereas the EAS rewards take approximately 40,000 episodes to converge. This slower convergence in the EAS can be attributed to the added complexity of its reward function, which incorporates not only the displacement in the target direction but also an energy penalty. We set a sufficiently large number of episodes ($N_E=100,000$) to ensure convergence of the reward function while maintaining a manageable training time. Similarly, a sufficiently large number of action steps per episode ($N_s=200$) is set to yield a high success rate of navigation while keeping training time minimal (see Supplemental Material~\cite{supplement} for more details on the effect of $N_s$ on the success rate.)



\section{Results and Discussion} \label{sec:results}

\subsection{Stroke patterns and motion dynamics}

\begin{figure*}[t]
\centering
 	\includegraphics[scale=0.8]{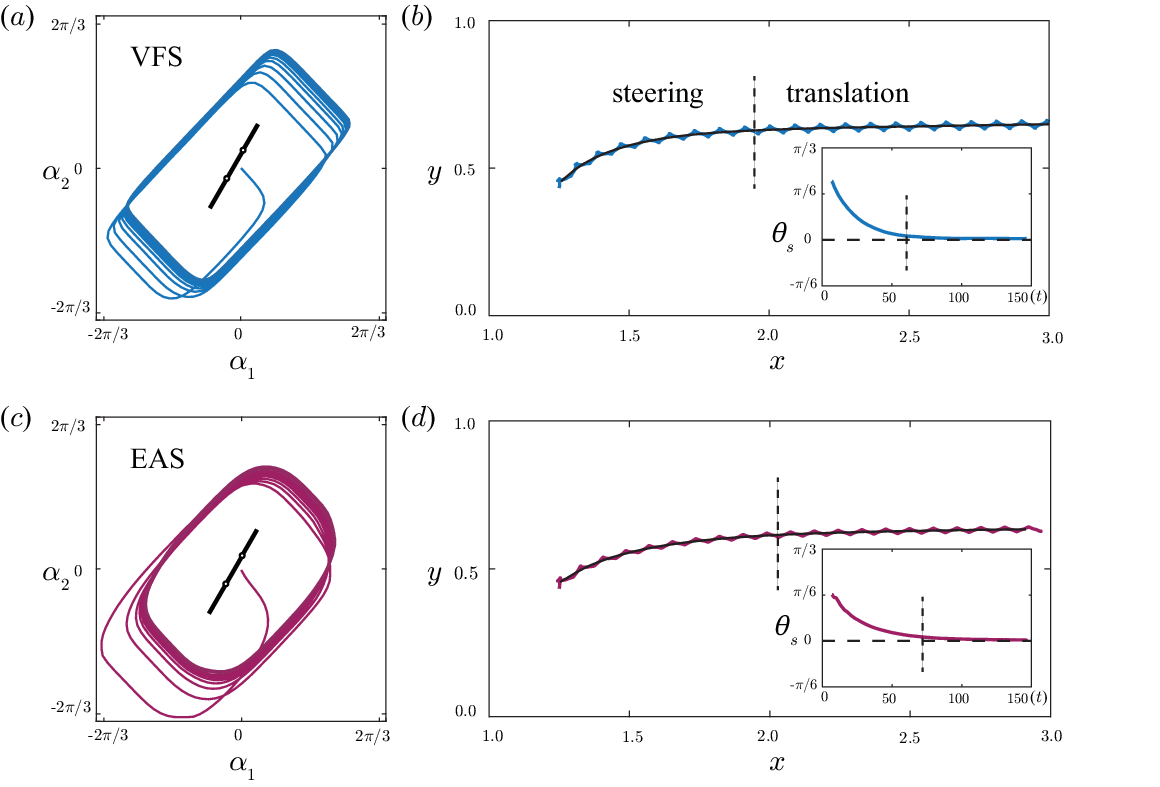}
 	\caption{\footnotesize Learning to swim along $\theta_T = 0$ with VFS and EAS. $(a, c)$: Stroke patterns in the phase plane. $(b, d)$: Trajectories of the geometric centroid and the smoothed path. The initial state is set as $\x_1 = (1,0)$, $\theta_1 = \theta_2 = \theta_3 = \pi/3$. The insets in $(b, d)$ display the evolution of the swimmer's averaged orientation, $\theta_s$, over time. In $(b, d)$, the black lines represent the smoothed path of the swimmer's motion, with the black dashed line used to distinguish the steering and translation stages. The blue lines indicate the VFS results, while the purple lines show the EAS results.} 
 	\label{fig:trajectory}
\end{figure*}

In Fig.~\ref{fig:trajectory}, we illustrate the swimming trajectories based on the VFS and EAS. The initial configuration of the swimmer is set as $\x_1 = (1,0)$ and $\theta_1 = \theta_2 = \theta_3 = \pi/3$. The swimmer, following both strategies, is allowed to move for 1500 steps. The trajectories of the stroke patterns in the phase plane are shown in Fig.~\ref{fig:trajectory}$(a,c)$, while the corresponding trajectories of the geometric centroid of the swimmer in the physical space are shown in Fig.~\ref{fig:trajectory}$(b,d)$.

We observe that in both cases, the swimmer successfully achieves targeted navigation and swims horizontally. The trajectories can be divided into two stages: steering and translation. The steering is the process of the swimmer adjusting its direction, while the translation reflects the swimmer moving steadily along a given direction. In the phase space shown in Fig.~\ref{fig:trajectory}$(a,c)$, the phase point circles clockwise, with the swirling center gradually approaching the origin of the phase plane. The stroke pattern eventually converges to a symmetric closed loop, indicating straight locomotion. In the physical space shown in Fig.~\ref{fig:trajectory}$(b,d)$, the swimmer gradually turns clockwise and ultimately swims horizontally. The transient process represents the steering stage, while the converged straight motion represents the translation stage. The converged stroke patterns of VFS and EAS are visibly different. The VFS trajectory in the phase space is more rectangular, while the EAS trajectory is more rounded.

Since the position of the swimmer's centroid oscillates during motion, we need to define an averaged orientation to establish a criterion for convergence. Observing that one period of the swimmer's motion contains about 70 steps, we choose to smooth the trajectories by averaging over this period. Specifically, for each step \( i \) (where \( i > 35 \)), we calculate the average position by considering the positions from 35 steps before to 35 steps after step \( i \). This means we average the positions from step \( i - 35 \) to step \( i + 35 \), effectively smoothing over one full period of motion. The smoothed path is shown as the black solid lines in Fig.~\ref{fig:trajectory}$(b,d)$. Next, we calculate the slope angle \( \theta_s \) of the smoothed path to determine the averaged orientation of the swimmer. To do this, we compute the finite differences between consecutive smoothed positions to obtain the local slope at each point.   By analyzing \( \theta_s \), we can assess how effectively the swimmer is aligning its motion with the desired target direction, thereby establishing a criterion for convergence. In the insets of Fig.~\ref{fig:trajectory}$(b,d)$, we show the convergence of the averaged orientation, $\theta_s$. In addition, we use $\theta_s$ to precisely distinguish between the steering and translation stages. If $|\theta_s| > 2.5^{\circ}$, the trajectory segment is classified as steering; otherwise, it is classified as translation. Based on this classification, we use a dashed line to separate the two stages.

\subsection{Swimming speed and efficiency}

\begin{figure*}[t]
\centering
 	\includegraphics[scale=0.8]{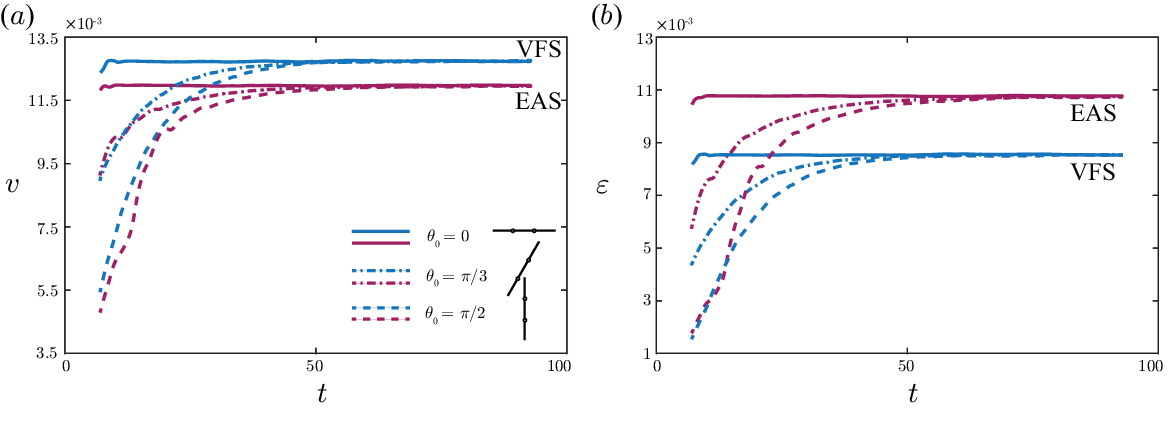}
 	\caption{\footnotesize Velocity along the target direction $(a)$ and swimming efficiency $(b)$ over time. Velocity and efficiency are calculated based on the smoothed paths. In each case, straight initial configurations with $\theta_2 = 0$ (solid line), $\pi/3$ (dash-dotted line), and $\pi/2$ (dashed line) are considered. } 
 	\label{fig:VOvsEOv}
\end{figure*}

Building upon the results that demonstrate both strategies effectively achieve targeted navigation, we proceed to quantitatively distinguish the VFS and EAS. By calculating the swimming speed along the target direction and the swimming efficiency during the steering and translation stages—based on the smoothed motion—we quantify the differences between the two strategies.

In Fig.~\ref{fig:VOvsEOv}$(a)$, we demonstrate that the translation stage is independent of the initial configurations. We simulate the dynamics resulting from the VFS and EAS with initial configurations $\theta_2 = \theta_0 = 0, \pi/3, \pi/2$ and $\alpha_1 = \alpha_2 = 0$. In both VFS and EAS, the horizontal speed, denoted $v$,  converges to the same value. For the VFS, the horizontal speed converges to approximately $0.01284$, while for the EAS, the steady speed is slightly slower at about $0.01176$.

To evaluate the swimming efficiency, we adopt a definition by Lighthill \cite{lighthill1975mathematical} and Purcell \cite{purcell1997efficiency}. At a given time, we calculate the rate of work done by the swimmer on the fluid, denoted as $\Phi(t)$, using Eq.~\ref{eq:energy}. As a reference motion, we consider towing the swimmer in its straightened configuration $(\alpha_1 = \alpha_2 = 0, \ \theta_2 = 0)$ along the horizontal direction at velocity $v(t)$. The rate of work for the towing problem is calculated as:
\begin{align}\label{eq:energyw_0}
    \Phi_0 =\gamma  v^2.
\end{align}
 The swimming efficiency, $\varepsilon$, is then defined as the ratio of the rates of work: 
\begin{align}\label{eq:varepsilon}
    \varepsilon=\frac{\Phi_0}{\Phi}.
\end{align}
By calculating the swimming efficiency, we observe that in both VFS and EAS, the efficiency criterion $\varepsilon$ converges to consistent values despite different initial configurations. According to Fig.~\ref{fig:VOvsEOv}$(b)$, for the VFS, the efficiency converges to approximately $0.854\%$, while for the EAS, a higher efficiency of about $1.077\%$ is achieved. These results demonstrate that, regardless of the initial configuration, both strategies converge to their respective steady efficiency levels.

\begin{figure*}[t]
\centering
 	\includegraphics[scale=0.8]{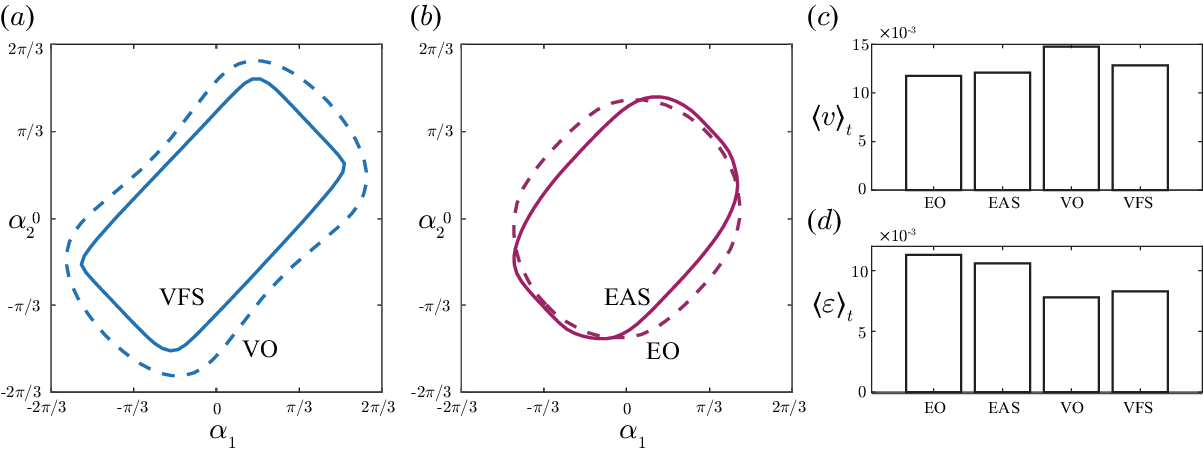}
 	\caption{\footnotesize Comparison between the models from optimizations and the strategies obtained through RL. $(a)$: Stroke patterns with VO (Velocity Optimal) and VFS (Velocity-Focused Strategy). $(b)$: Stroke patterns with EO (Efficiency Optimal) and EAS (Energy-Aware Strategy). $(c)$: Comparison of average velocity along the target direction for all four strategies. $(d)$: Comparison of average swimming efficiency for all four strategies. The results of VO and EO are reproduced from Ref.~\cite{tam2007optimal}.}  
 	\label{fig:AIvsTam}
\end{figure*}

During the translation stage, the swimmer begins to move steadily by repeating the same stroke pattern. In Fig~\ref{fig:AIvsTam}$(a,b)$, we plot the converged stroke patterns for both VFS and EAS in the phase space using solid lines. Tam \textit{et al.}~\cite{tam2007optimal} investigated the optimal stroke patterns for the three-link swimmer, focusing on two cases: velocity optimal (VO) and efficiency optimal (EO). They numerically optimized the periodic functions of $\alpha_1$ and $\alpha_2$ using gradient search. In Fig.~\ref{fig:AIvsTam}$(a,b)$, we reproduce the stroke patterns of VO and EO in \cite{tam2007optimal} using dashed lines. We compare these optimal stroke patterns with those obtained through our reinforcement learning approach. It is intriguing to observe that, although the RL-generated stroke patterns are not identical to the optimized patterns from \cite{tam2007optimal}, they exhibit similar features. For instance, the stroke pattern from the VFS shares similarities with the VO pattern, being more rectangular in the phase space. Meanwhile, the stroke pattern from the EAS resembles the EO pattern, appearing more rounded.

In Fig.~\ref{fig:AIvsTam}$(c,d)$, we further quantitatively compare the results from our RL strategies with those from \cite{tam2007optimal} by calculating the average velocity along the target direction $\langle v \rangle_t$ and the swimming efficiency over one stroke cycle $\langle \varepsilon \rangle_t$. Specifically, we consider four cases: the velocity-optimal (VO) and the efficiency-optimal (EO) stroke patterns from optimizations, the Velocity-Focused Strategy (VFS), and the Energy-Aware Strategy (EAS) from RL. 

For the average velocity, the VO achieves the highest value, followed by the VFS, EAS, and EO. Quantitatively, the VFS achieves over 80\% of the average velocity of the VO, indicating that the RL-generated VFS closely approximates the velocity performance of the optimal stroke pattern. In terms of swimming efficiency, the EO from optimization attains the highest efficiency, followed by the EAS, VFS, and VO. Notably, the EAS again captures over 80\% of the efficiency achieved by the EO, which demonstrates that the RL-generated EAS effectively balances energy consumption while maintaining reasonable propulsion.

These results highlight that, although the stroke patterns obtained through RL are not identical to the optimal ones, they exhibit similar features and achieve comparable performance levels. This underscores the capability of RL in developing effective stroke patterns that align with specific objectives, such as maximizing velocity or efficiency, without explicitly programming these optimal solutions. Overall, the RL approach demonstrates a strong ability to capture key characteristics of optimal swimming gaits identified by traditional optimization methods.

\subsection{Complex navigation tasks}

\begin{figure*}[t]
\centering
 	\includegraphics[scale=0.8]{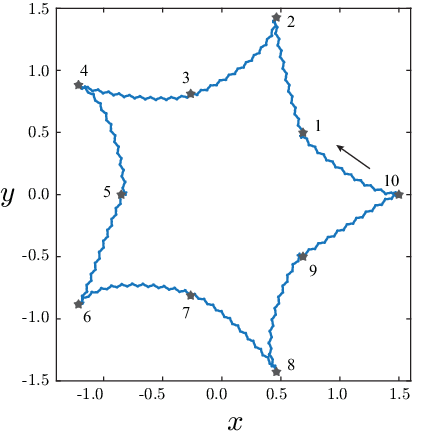}
 	\caption{\footnotesize Three-link swimmer traces a star-shaped trajectory. The trajectory of the swimmer's geometric centroid is represented by blue lines. Initialized in a straight configuration with $\theta_1 = \theta_2 = \theta_3 = 0$, the swimmer is provided with a sequence of target points (1–10), where it chases one target point (grey stars) at a time. The black arrows indicate the intended direction of the swimmer's movement.}
 	\label{fig:navigatestar}
\end{figure*}

Finally, we showcase the swimmer's capability to trace complex paths and navigate towards moving targets. In Fig.~\ref{fig:navigatestar}, we task the swimmer with tracing a star-shaped trajectory. Notably, the hydrodynamic calculations required to design the stroke patterns for such complex paths can become intractable as the complexity increases. Here, rather than explicitly programming the swimmer's stroke patterns, we only select target points ($\x_{T_i}, i = 1, 2, \dots, 10$) as landmarks and require the swimmer to navigate using its own strategy. The target direction at time step $k+1$ is given by $\theta_{T_{k+1}}=\arg(\x_{T_i}-\x_{c_k})$. Starting from the initial state $\mathbf{x}_1 = (1, 0)$ with link orientations $\theta_1 = \theta_2 = \theta_3 = 0$, the swimmer, equipped with the VFS model, is assigned the next target point $\x_{T_{i+1}}$ once its centroid is within a certain threshold (set to 0.001 here) from $\x_{T_i}$. The navigation strategy enables the swimmer to adjust its swimming gaits to navigate several wide (e.g., around point 3) and sharp angles (e.g., around point 4)  in tracing the star-shaped trajectory (see Supplementary Movie 1~\cite{supplement}).

\begin{figure*}[t]
\centering
 	\includegraphics[scale=0.8]{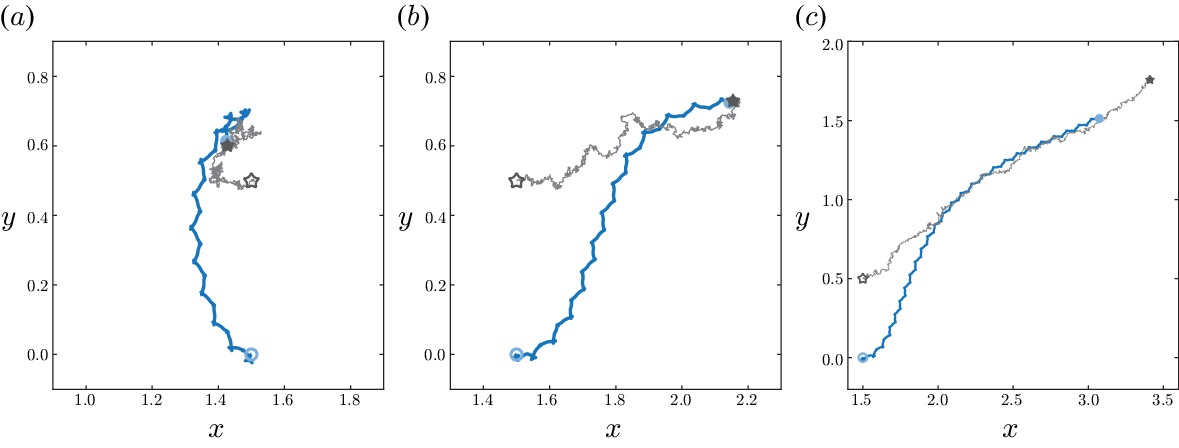}
 	\caption{\footnotesize Three-link swimmer (VFS, blue dots) navigates towards diffusing targets (grey stars) with different speeds.  ($a$): $v_T/v_m=0$. ($b$): $v_T/v_m=0.5$. ($c$): $v_T/v_m=1$; $v_T$ is the target's speed, and  $v_m$ is the  maximum speed achieved by the VFS. The trajectory of the swimmer's geometric centroid is shown in blue lines, and the trajectory of the target is shown in grey lines. The swimmer is initialized as a straight shape with $\theta_1 = \theta_2 = \theta_3 = 0$, and the target is oriented at $30^\circ$ relative to the horizontal axis. The diffusivity is set to $D=5\times10^{-5}$. }
 	\label{fig:navigatediffusingtargets}
\end{figure*}

Next, we demonstrate the RL-powered swimmer’s capability to navigate toward a dynamic target, characterized by its position $\mathbf{x}_T$, orientation $\mathbf{p}_T$, and intrinsic speed $v_T$. In addition, we consider scenarios where the target's movement is influenced by random fluctuations due to Brownian motion, characterized by a diffusivity $D$. 
This target undergoes purely translational diffusion in two dimensions, described by independent Brownian motions in the $x$- and $y$-directions. Specifically, each action step satisfies $\langle \delta x^2 \rangle=\langle \delta y^2 \rangle=2D\delta t$, where $\delta t$ denotes the duration of an action step, and $\delta x$ and $\delta y$ are the displacement in the $x$- and $y$-directions within one action step.
This combination of directed movement and random motion of the target introduces additional complexity to the swimmer's navigation task. We note that all these quantities---position, orientation, speed, and diffusivity---are nondimensionalized using the characteristic length, time, and force scales defined earlier in \S\ref{sec:model}.

In Fig.~\ref{fig:navigatediffusingtargets}, we present three scenarios where the swimmer navigates toward moving targets with different intrinsic speeds. The swimmer utilizes the VFS to adjust its stroke patterns based on the current observed direction of the moving target relative to its own position.Navigation is achieved by continuously sensing the target's location and adapting its movements to minimize the distance between the swimmer and the target. In the simulations, the swimmer's initial state is set as $\mathbf{x}_1 = (1, 0)$ with link orientations $\theta_1 = \theta_2 = \theta_3 = 0$. The target starts from the initial position $\mathbf{x}_T = (1.5, 0.5)$ and has an initial orientation of $30^\circ$ relative to the horizontal axis. The diffusivity of the target is set to $D = 5 \times 10^{-5}$. We consider targets with three different intrinsic speeds: $v_T / v_m = 0$, $0.5$, and $1$, where $v_m$ denotes the maximum speed achieved by the VFS. We define capture as the event when the distance between the swimmer and the target becomes less than a predefined threshold of 0.001.  Once the swimmer is within this distance of the target, it is considered to have successfully captured the target. 

In Fig.~\ref{fig:navigatediffusingtargets}$(a)$, we present the scenario where the swimmer (represented by the blue dot) navigate towards the target (grey star) undergoing pure Brownian motion (i.e. $v_T = 0$). Despite the target's random motion, the swimmer effectively adjusts its motion based on the observed direction of the target relative to its own position and navigates toward the moving target. We observe that the swimmer's centroid follows a relatively smooth trajectory compared with the randomly fluctuating path of the target. The swimmer eventually captures the moving target (see Supplementary Movie 2~\cite{supplement}). When the target has an intrinsic speed that is half that of the swimmer (i.e., $v_T = v_m/2$) in addition to its random motion, the swimmer is still able to continuously adapt its stroke patterns to pursue and successfully capture the moving target (see Supplementary Movie 3~\cite{supplement}). In Fig.~\ref{fig:navigatediffusingtargets}(c), we push the limits further by examining the scenario where the moving target's intrinsic speed is increased to match that of the swimmer (i.e., $v_T =v_m$). Under this challenging condition, the swimmer is unable to capture the target but is still able to closely follow its trajectory (see Supplementary Movie 4~\cite{supplement}). Taken together, these results demonstrate the capability of the RL-powered swimmer to navigate toward a target moving at a significant fraction of its own speed.

\section{Concluding Remarks}\label{sec:conclusion}

In this work, we presented a reinforcement learning (RL) approach to enable the navigation of a three-link swimmer at low-Reynolds numbers. While a prior study demonstrated limited locomotion of a three-link swimmer with discrete action spaces \cite{qin2023reinforcement}, the deep RL-powered swimmer presented here leverages continuous action spaces to learn complex stroke patterns for effective swimming and navigation toward a target direction. We examined how different reward functions -- one that rewards only the swimmer's velocity toward the target and another that also accounts for energy consumption -- lead to the development of distinct stroke patterns. We note that energetic cost has been incorporated into the reward function in previous predator-prey contexts \cite{zhu2022optimizing}. In contrast, our work focuses on the propulsion performance of a widely studied three-link swimmer, enabling direct benchmarking of RL-derived strategies against those obtained from prior optimization-based approaches. With different reward functions, we observed that the RL-derived stroke patterns exhibit qualitative features similar to the optimal solutions identified in previous optimization studies \cite{tam2007optimal}. Quantitatively, the strategies developed by RL are at least 80\% as effective as the optimal solutions in terms of both propulsion velocity and energetic efficiency. The performance gap may be attributed to the fundamental difference in methodology: prior optimization-based approaches typically impose a single-period optimization of stroke kinematics and explicitly search for an optimal periodic gait under well-defined parameters, the RL framework applies no such constraints \textit{a priori}. Instead, the RL agent is free to discover any control strategy that achieves forward motion, without assuming periodicity or a fixed stroke duration. While additional stroke constraints may be incorporated into the RL framework, such approaches inherently impose a preferred structure on the solution. In contrast, the simpler reward functions used here allow the RL agent to more autonomously develop its own strategies, providing a flexible alternative for complex scenarios where effective stroke patterns are not well understood or where the setup may be dynamically changing. Lastly, we demonstrated the swimmer's ability to autonomously adapt its stroke patterns to navigate in any target direction, enabling it to trace complex trajectories and pursue moving targets (e.g., mimicking swimming bacteria or circulating tumor cells). These capabilities serve as proof of concept for scenarios relevant to potential biomedical applications.



We remark on several limitations of the current study and discuss potential directions for future research. First, we use a three-link swimmer here as a simple example to demonstrate the RL approach. We anticipate that increasing the degrees of freedom by incorporating additional links will enable a multi-link swimmer to perform more complex maneuvers and further enhance propulsion performance. Second, in demonstrating the swimmer's ability to pursue a moving target, we neglect the hydrodynamic interactions between the swimmer and the target. Incorporating these interactions in future work could reveal new features in the strategies identified by RL. Lastly, the presence of obstacles or flow perturbations in complex biological environments would also impact the swimmer's navigation. Addressing these challenges would pave the way for developing intelligent microswimmers with more robust navigation capabilities.

\section*{Acknowledgments}

Y.~Lai acknowledges partial support from the National Natural Science Foundation of China (NSFC) through the Fundamental Research Project for Undergraduates (Grant No.~123B1034).  
O.~S.~Pak acknowledges partial support from the National Science Foundation (NSF) under Grant Nos.~CBET-2323046 and CBET-2419945.  
Y.~Man acknowledges partial support from the NSFC under Grant No.~12372258.

\end{document}


\preprint{}

\title{\textnormal{Supplemental Material for} \\ Navigation of a Three-Link Microswimmer via Deep Reinforcement Learning}


\author{Yuyang Lai}
\affiliation{Department of Mechanics and Engineering Science at College of Engineering, Beijing 100871, PR China}

\author{Sina Heydari}
\affiliation{Department of Mechanical Engineering, Santa Clara University, Santa Clara, CA 95053, USA}

\author{On Shun Pak}
\email[Email address for correspondence:]{opak@scu.edu}
\affiliation{Department of Mechanical Engineering, Santa Clara University, Santa Clara, CA 95053, USA}
\affiliation{Department of Applied Mathematics, Santa Clara University, Santa Clara, CA 95053, USA}

\author{Yi Man}
\email[Email address for correspondence:]{yiman@pku.edu.cn}
\affiliation{Department of Mechanics and Engineering Science at College of Engineering, Beijing 100871, PR China}



\date{\today}


\maketitle

The supplementary material consists of three sections. The first section details the dynamic model of the three-link swimmer, including the derivation of its equations of motion. The second section explains the PPO framework for training the swimmer's control policy, outlining the Actor-Critic network and the reward structure. The third section examines the impact of training parameters on the swimmer's performance.

\section{Equations of motion}

\begin{figure*}[t]
\centering
 	\includegraphics[scale=0.8]{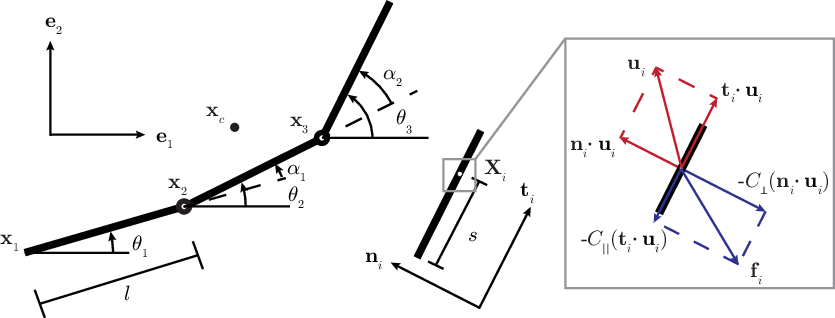}
 	\caption{\footnotesize Model of three-link swimmer. It consists of three rigid links of equal length, connected by two hinges, allowing rotation to adjust the relative angles $\alpha_i$ $(i = 1, 2)$. With 5 degrees of freedom, the coordinates of the three-link swimmer can be determined by the coordinates of one end $x_1$ and the orientations of three links, $\theta_1, \theta_2, \theta_3$. We adopt resistive force theory to derive the force acting on the swimmer.} 
 	\label{fig:complex3link}
\end{figure*}

We consider a three-link swimmer with total length $L$, where each link has a radius $a$ and a length $l = L/3$ (see Fig.~\ref{fig:complex3link}). To derive the dominant dynamic equations of the three-link swimmer, we begin by computing the force acting on any point along link $i$. The position of a point on link $i$ is given by $\X_i = \x_i + s\t_i$, where $s$ denotes the distance from the left end of the link, and $\t_i$ is the unit tangent vector along the link. The local velocity at $\X_i$ can then be expressed as:
\begin{align}\label{eq:localvelocity}
\u_i = \dot\x_i+s\dot\theta_i \n_i, 
\end{align}
where $\n_i$ represents the unit vector normal to link $i$. Based on resistive force theory, the local hydrodynamic force is directly proportional to the local velocity \cite{lighthill1975mathematical}. As a result, the local force is computed as:
\begin{align}\label{eq:localforce}
    \f_i=-\left(C_{\parallel}\t_i\t_i+C_\perp\n_i\n_i\right)\cdot\u_i.
\end{align}
Here $C_\parallel=2\pi \mu/\left[\ln\left(L/a\right)-1/2\right]$ and $C_\perp=4\pi \mu/\left[\ln\left(L/a\right)+1/2\right]$ are the drag coefficients \cite{lighthill1975mathematical}, with $\mu$ denoting the dynamic viscosity of the fluid. By substituting Eq.~\eqref{eq:localvelocity} into Eq.~\eqref{eq:localforce}, the resulting hydrodynamic force is:
\begin{align}
    \f_i=-\left(C_{\parallel}\t_i\t_i+C_\perp\n_i\n_i\right)\cdot\dot\x_i-C_\perp s\dot\theta_i \n_i.
\end{align}

By integrating along link $i$, the total force is expressed as:
\begin{align}
    \F_i&=\int_0^l\f_i\:\mathrm{d}s
    \notag
    \\ & =-l\left(C_{\parallel}\t_i\t_i+C_\perp\n_i\n_i\right)\cdot\dot\x_i-\frac{1}{2}C_\perp l^2\dot\theta_i \n_i
    \notag
    \\ & =-C_{\parallel}l\left(\dot\x_i\cdot\t_i\right)\t_i-C_\perp\left[l\left(\dot\x_i\cdot\n_i\right)+\frac{1}{2}l^2\dot\theta_i\right]\n_i.
\end{align}
The components of the force are calculated as follows:
\begin{subeqnarray}\label{eq:totalforcexy}
    F_{ix}&=&l\left[-\left(C_\parallel\cos^{2}{\theta_{i}}+C_\perp\sin^{2}{\theta_{i}}\right)\dot{x}_{i}+\left(C_\perp-C_\parallel\right)\sin{\theta_{i}}\cos{\theta_{i}}\dot{y}_{i}+\frac{1}{2}C_\perp l\sin{\theta_{i}}\dot{\theta}_{i}\right], 	\\
    F_{iy}&=&l\left[\left(C_\perp-C_\parallel\right)\sin{\theta_{i}}\cos{\theta_{i}}\dot{x}_{i}-\left(C_\parallel\sin^{2}{\theta_{i}}+C_\perp\cos^{2}{\theta_{i}}\right)\dot{y}_{i}-\frac{1}{2}C_\perp l\cos{\theta_{i}}\dot{\theta}_{i}\right].
\end{subeqnarray}

Referring to Eq.~\eqref{eq:localforce}, the total hydrodynamic torque  can be calculated as:
\begin{align}
    \M_i&=\int_0^l\X_i\times\f_i\mathrm{d}s
    \notag
    \\ &=\int_0^l\left(\x_i+s\t_i\right)\times\f_i\mathrm{d}s
    \notag
    \\ &=\x_i\times\F_i+\int_0^ls\t_i\times\f_i\mathrm{d}s.
\end{align}
The cross product $\t_i\times\f_i$ is computed as follows:
\begin{align}
    \t_i\times\f_i&=-\t_i\times\left[\left(C_{\parallel}\t_i\t_i+C_\perp\n_i\n_i\right)\cdot\dot\x_i\right]-C_\perp s\dot\theta_i\e_3
    \notag
    \\ &=-C_\perp\left[\left(\dot\x_i\cdot\n_i\right)+s\dot\theta_i\right]\e_3.
\end{align}
Here, $\e_3=\e_1\times\e_2$ denotes the unit vector normal to the plane. The total hydrodynamic torque then becomes:
\begin{align}\label{totaltorque}
    \M_i=\x_i\times\F_i-C_\perp l^2\left[\frac{1}{2}\left(\dot\x_i\cdot\n_i\right)+\frac{1}{3}l\dot\theta_i\right]\e_3.
\end{align}
Finally, the resultant hydrodynamic torque per link, $M_i = \M_i\cdot\e_3$, is expressed as:
\begin{align}
    M_i=& 
    \notag
    \\ &
    l\left[\frac{C_\perp}{2}l\sin{\theta_i}+\frac{1}{2}\left(C_\perp-C_\parallel\right)x_i\sin{2\theta_i}+\frac{1}{2}\left(C_\perp+C_\parallel\right) y_i-\frac{1}{2}\left(C_\perp-C_\parallel\right)y_i\cos{2\theta_i} \right]\dot x_i
    \notag
    \\ &-l\left[\frac{C_\perp}{2}l\cos{\theta_i}+\frac{1}{2}\left(C_\perp-C_\parallel\right)y_i\sin{2\theta_i}+\frac{1}{2}\left(C_\perp+C_\parallel\right)x_i+\frac{1}{2}\left(C_\perp-C_\parallel\right)x_i\cos{2\theta_i} \right]\dot y_i
    \notag
    \\ &-C_\perp l^2\left[\frac{1}{3}l+\frac{1}{2}x_i\cos{\theta_i}+\frac{1}{2}y_i\sin{\theta_i}\right]\dot\theta_i.
\end{align}

According to the absence of inertia in low-Reynolds number swimming, the total hydrodynamic force and torque acting on the swimmer must be zero. Thus, we have:
\begin{align}\label{eq:forcebalance}
\sum_{i=1}^{3}F_{ix}=0, \quad \sum_{i=1}^{3}F_{iy}=0, \quad \sum_{i=1}^{3}M_i=0.
\end{align}

Moreover, the motion of the swimmer is subject to geometric constraints (where $i=1,2$) as follows:
\begin{subeqnarray}\label{eq:gconstraints}
x_{i+1}-x_{i}=l\cos{\theta_{i}}, 	\\
y_{i+1}-y_{i}=l\sin{\theta_{i}},  \\
\theta_{i+1}-\theta_i=\alpha_i.
\end{subeqnarray}
Taking the derivative with respect to $t$, we derive the kinematic constraints as:
\begin{subeqnarray}\label{eq:constraints}
\dot x_{i+1}-\dot x_{i}=-l\dot \theta\sin{\theta_{i}}, 	\\
\dot y_{i+1}-\dot y_{i}=l\dot \theta\cos{\theta_{i}},  \\
\dot\theta_{i+1}-\dot\theta_i=\dot\alpha_i.
\end{subeqnarray}

In presenting our results, all lengths are scaled by the total length of the swimmer, $L$. A characteristic time scale, $T_0$, is assumed, which corresponds to the actuation rate of the angle between neighboring links. The associated force scale is defined as $C_\perp L^2/T_0$. As a result, the dimensionless quantities are defined as $\x_i = L \overline{\x}_i$, $\overline{\dot{\alpha_j}} = T_0 \dot{\alpha_j}$, and $\gamma = C_\parallel / C_\perp$, where $i = 1,2,3$ and $j = 1,2$. In this study, we consider a slender swimmer ($a \ll L$) with $\gamma = 1/2$. For simplicity, we omit the overbars hereafter and use only dimensionless quantities. Combining Eqs.~\eqref{eq:forcebalance} and \eqref{eq:constraints}, we derive a complete description of the swimmer's motion, expressed as a system of linear equations:
\begin{align}\label{eq:motion}
    \boldsymbol{H}(\boldsymbol{X},\boldsymbol{Y},\boldsymbol{\Theta})
\begin{pmatrix}
  \dot{\boldsymbol{X}} \\
  \dot{\boldsymbol{Y}} \\
  \dot{\boldsymbol{\Theta}}
\end{pmatrix}=\boldsymbol{q},
\end{align}
where $\boldsymbol{X}=[x_1,x_2,x_3]^\top$, $\boldsymbol{Y}=[y_1,y_2,y_3]^\top$, and $\boldsymbol{\Theta}=[\theta_1,\theta_2,\theta_3]^\top$, while $\dot{\boldsymbol{X}}$, $\dot{\boldsymbol{Y}}$, and $\dot{\boldsymbol{\Theta}}$ represent their respective time derivatives with respect to $t$. Most elements of $\boldsymbol{q}$ are nearly zero, except for two non-zero components, $\dot{\alpha}_1$ and $\dot{\alpha}_2$.

Then, the nontrivial components of the matrix $\boldsymbol{H}(\boldsymbol{X},\boldsymbol{Y},\boldsymbol{\Theta})$ are as follows (for $i=1,2,3$):
\begin{equation}
\begin{split}
    & H_{1i} =  -\frac{ 1}{3}\left(\gamma\cos^{2}{\theta_i}+\sin^{2}{\theta_i}\right), \quad H_{1,i+3} = \frac{ 1}{6}\left(1-\gamma\right)\sin{2\theta_i}, \quad H_{1,i+6} = \frac{1}{18}\sin{\theta_i}, \\
    & H_{2i} = \frac{1}{6}\left(1-\gamma\right)\sin{2\theta_i}, \quad  H_{2,i+3} = -\frac{ 1}{3}\left(\cos^{2}{\theta_i}+\gamma\sin^{2}{\theta_i}\right), \quad H_{2,i+6} =  -\frac{ 1}{18}\cos{\theta_i}, \\
    & H_{3i}  =  \frac{1}{18}\sin{\theta_i}+\frac{1}{6}\left(1-\gamma\right)x_i\sin{2\theta_i} +\frac{1}{6}\left(1+\gamma\right) y_i-\frac{1}{6}\left(1-\gamma\right)y_i\cos{2\theta_i} , \\
    & H_{3,i+3} =  -\left[\frac{1}{18}\cos{\theta_i}+\frac{1}{6}\left(1-\gamma\right)y_i\sin{2\theta_i} +\frac{1}{6}\left(1+\gamma\right) x_i+\frac{1}{6}\left(1-\gamma\right)x_i\cos{2\theta_i} \right], \\
    & H_{3,i+6} =  -\left[\frac{1}{81}+\frac{1}{18}x_i\cos{\theta_i}+\frac{1}{18}y_i\sin{\theta_i} \right], \\
    &H_{41}=H_{52}=H_{64}=H_{75}=H_{87}=H_{98}=-1, \quad H_{42}=H_{53}=H_{65}=H_{76}=H_{88}=H_{99}=1,\\
    & H_{47} = \frac{1}{3}\sin\theta_1, \quad H_{58} =\frac{1}{3}\sin\theta_2,\quad H_{67}=-\frac{1}{3}\cos\theta_1, \quad  H_{78}=-\frac{1}{3}\cos\theta_2. 
\end{split}
\end{equation}

The nontrivial components of the dimensionless vector $\boldsymbol{q}$ are expressed as:
\begin{align}
 q_{8} = \dot{\alpha}_1, \quad q_{9} = \dot{\alpha}_2.
\end{align}

\section{Proximal Policy Optimization (PPO) algorithm}

\begin{figure*}[t]
\centering
 	\includegraphics[scale=0.8]{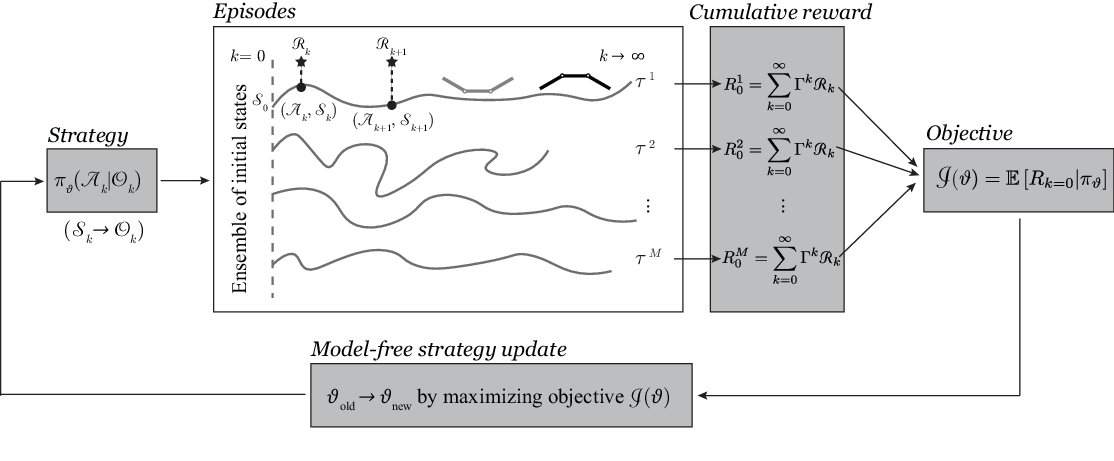}
 	\caption{\footnotesize RL update cycle for the targeted navigation task: RL aims to learn a strategy $\pi_{\vartheta}\left(\mathcal{A}_k|\mathcal{O}_k\right)$, parameterized by $\vartheta$, that maximizes an objective function  $\mathcal{J}\left(\vartheta\right)$. Repeated interactions with the environment drive this learning process. Starting with an initial strategy—potentially random—the RL agent uses a noisy version of this strategy to explore the environment, collecting an ensemble of trajectories. Based on these trajectories, the strategy is iteratively updated.} 
 	\label{fig:complexmodel}
\end{figure*}

We employ the Proximal Policy Optimization (PPO) algorithm with an Actor-Critic neural network agent for reinforcement learning (RL) training. We use $k$ to represent the ordinal number of time steps. The Actor network represents a stochastic control strategy, $\pi_{\vartheta}\left(\mathcal{A}_k|\mathcal{O}_k\right)$, which generates an action $\mathcal{A}_k$ based on the observation $\mathcal{O}_k$, following a Gaussian distribution. The Critic network computes the expected return function $V_\varphi = V_\varphi\left(\mathcal{O}\right)$, which estimates the expected return when the agent starts from observation $\mathcal{O}$ and follows the strategy $\pi_{\vartheta}$ to take actions. The parameters of the Actor network are denoted by $\vartheta$. The Critic network computes the expected return function $V_\varphi$, given the agent's observation $\mathcal{O}_k$ and its control strategy $\pi_{\vartheta}$. The parameters of the Critic network are denoted by $\varphi$. PPO is a policy gradient method that ensures efficient learning and robust performance by limiting the change allowed to the policy during each update.

The training process is divided into $N_E$ episodes, each consisting of $N_s$ action steps. In this study, we set $N_E=100,000$ to provide a reasonable total duration for computation and ensure convergence. At the same time, $N_s=200$ is chosen to ensure that the swimmer performs well in both turning maneuvers and stable straight swimming. To promote full exploration of the observation space, the swimmer starts from a random initial state, $\mathcal{S}_0$.

At each time step $k$, the agent receives the current observation $\mathcal{O}_k$ and samples an action $\mathcal{A}_k$ according to the strategy $\pi_{\vartheta}$. Using the dynamics in Eq.~\ref{eq:motion}, the agent calculates the next state $\mathcal{S}_{k+1}$ through time integration over a time step $\delta t$, and calculates the corresponding reward $\mathcal{R}_k$. The next observation $\mathcal{O}_{k+1}$, derived from $\mathcal{S}_{k+1}$, is provided to the agent to inform the subsequent action. This interactive process forms a trajectory $\tau$, which consists of states, observations, actions, and rewards: $\tau=\left(\mathcal{S}_0, \mathcal{O}_0, \mathcal{A}_0, \mathcal{R}_0, ..., \mathcal{S}_{N_s}, \mathcal{O}_{N_s}, \mathcal{A}_{N_s}, \mathcal{R}_{N_s}\right)$. The agent collects and stores these trajectories for subsequent updates.

After completing a fixed number of episodes, $M$ (leading to a total of $N_u=MN_s$ time steps), the agent updates the Actor network parameter,$\vartheta$, to maximize the RL objective $\mathcal{J}(\vartheta)$. This objective is derived from collected trajectories. As depicted in Fig.~\ref{fig:complexmodel}, $M$ trajectories, $\tau^1, \tau^2, \dots, \tau^M$, are collected during each update cycle, with the states, observations, actions, and rewards stored in the corresponding lists: states $\mathcal{S}_{N_u \times 5}$, observations $\mathcal{O}_{N_u \times 4}$, actions $\mathcal{A}_{N_u \times 2}$, and rewards $\mathcal{R}_{N_u \times 1}$. The list of rewards, $\mathcal{R}_{N_u \times 1}$, is then used to compute the cumulative reward list, $R_0^{M \times 1}$, where the cumulative reward is given by $R_{k}=\sum_{k'=k}^{\infty} \Gamma^{k'-k}\mathcal{R}_{k'}$. Here, $\Gamma \in \left[0,1\right]$ represents the discount factor, determining the trade-off between immediate and future rewards. A large discount factor, $\Gamma = 0.99$, is chosen to focus on long-term rewards. During each update cycle, the agent calculates the expected cumulative reward $\mathcal{J}(\vartheta) = \mathbb{E}[R_{k=0} | \pi_{\vartheta}]$ from the cumulative reward list $R_0^{M \times 1}$, using it as the RL objective, and derives an optimal strategy by maximizing this value.

The parameters $N_u$ and $M$ reflect the amount of data collected during each update cycle: larger values provide more thorough sampling and increase the overall training time. In this study, we set $N_u = 8,000$ and $M = 40$ to ensure the swimmer's performance is optimized through exhaustive exploration.

As an advanced policy gradient method, PPO, specifically the clipped PPO algorithm, estimates the objective $\mathcal{J}(\vartheta)$ by clipping the probability ratio $r(\vartheta)$ within the range $\left[1-\epsilon, 1+\epsilon\right]$, and multiplying it by the advantage function list $A_{N_u \times 1}$. The clipping constant, $\epsilon = 0.2$, specifies the maximum deviation allowed for the updated policy from the old one. In the $k$th time step, advantage function $A_k$ describes the relative benefit of taking an action $\mathcal{A}_k$ based on an observation $\mathcal{O}_k$ over a randomly selected action and is calculated by subtracting the expected returns function $V_k=V_{\varphi_{old}}\left(\mathcal{O}_k\right)$ from the cumulative reward $R_k$. The probability ratio $r\left(\vartheta\right)$ and the advantage function $A_k$ are defined as follows:
\begin{align}
   r\left(\vartheta\right) = \frac{\pi_{\vartheta}\left(\mathcal{A}_k|\mathcal{O}_k\right)_{N_u \times 1}}{\pi_{\vartheta_{old}}\left(\mathcal{A}_k|\mathcal{O}_k\right)_{N_u \times 1}}, \quad A_k = R_k-V_k.
\end{align}
The clipped surrogate advantage is then obtained as follows:
\begin{align}
    L^{CLIP}\left(\vartheta\right)=\mathbb{E}\{ \min\left[r(\vartheta)\cdot A_{N_u \times 1},clip(r(\vartheta),1-\epsilon,1+\epsilon)\cdot A_{N_u \times 1}\right] \}.  
\end{align}
Note that the Critic network's parameter, $\varphi$, does not appear explicitly in the clipped objective because $V_k$ is computed using the old expected returns function, $V_{\varphi_{old}}$. However, the parameter $\varphi$ is still updated to ensure the accuracy of the advantage function. To do this, we calculate the expected returns function $V_{N_u \times 1}$, and the cumulative reward list $R_{N_u \times 1}$. We then use the differences between these two lists to adjust the parameter of the Critic network, $\varphi$. Consequently, the expected returns loss function is defined as follows:
\begin{align}
    L^{ER}\left(\varphi\right)=\frac{1}{2}\mathbb{E}\left[\left(R_{N_u \times 1}-V_{N_u \times 1}\right)^2\right]. 
\end{align}

To promote deeper exploration by the swimmer, we define an entropy loss function for the actions, denoted as \( L^S \). This function increases the entropy of the actions during training, encouraging the agent to explore a wider range of actions and states. The entropy loss function \( L^S \) is defined as follows:
\begin{align}
    L^{S}\left(\vartheta\right)=\chi S\left[\pi_\vartheta\right],
\end{align}
where $ S\left[\pi_\vartheta\right] = -\sum_k \pi_\vartheta(\mathcal{A}_k|\mathcal{O}_k) \log \pi_\vartheta(\mathcal{A}_k|\mathcal{O}_k)$, and the constant $\chi = 0.01$ is a positive weight introduced to encourage exploration during training.

Finally, the surrogate loss function is defined as follows:
\begin{align}
    L\left(\vartheta,\varphi\right)=-L^{CLIP}\left(\vartheta\right)+L^{ER}\left(\varphi\right)-L^S\left(\vartheta\right).
\end{align}

We then update the parameters $\vartheta$ and $\varphi$ using a standard gradient descent algorithm, the Adam optimizer \cite{kingma2014adam}. This process determines the new strategy, $\pi_{\vartheta_{new}}$.

Our implementation of the PPO algorithm is summarized in Algorithm 1 and Algorithm 2. Here, $K$ represents the total number of epochs the Adam optimizer uses during the optimization process. These tables outline the specific steps of the PPO algorithm employed in this work, and we refer readers to classical references for further details \cite{schulman2017proximal}.

\begin{algorithm}
\caption{Environment}
\begin{algorithmic}[1]
\For{time step $k = 0, 1, \dots$}
    \If{$\bmod(k, N_s) = 0$}
        \State Reset state $\mathcal{S}_k$ and compute observation $\mathcal{O}_k$
    \EndIf
    \State Sample action $\mathcal{A}_k$ from strategy $\pi_{\vartheta_{old}}$
    \State Evaluate the next state $\mathcal{S}_{k+1}$ and reward $\mathcal{R}_k$ following the swimmer's hydrodynamics, then compute the next observation $\mathcal{O}_{k+1}$
    \If{$k = 0$ \textbf{or} $\bmod(k, N_u) \neq 0$}
        \State Append $\mathcal{O}_{k+1}$, $\mathcal{A}_k$, $\mathcal{R}_k$, and $\pi_\vartheta(\mathcal{A}_k|\mathcal{O}_k)$ to observation list $\mathcal{O}_{N_u \times 4}$, action list $\mathcal{A}_{N_u \times 2}$, reward list $\mathcal{R}_{N_u \times 1}$, and action sampling probability list $\pi_{\vartheta_{old}}(\mathcal{A}_k|\mathcal{O}_k)_{N_u \times 1}$
    \Else
        \State Update the agent according to Algorithm 2
    \EndIf
\EndFor
\end{algorithmic}
\end{algorithm}

\begin{algorithm}
\caption{Updating the Agent}
\begin{algorithmic}[1]
\State Initialize strategy parameters $\vartheta$ and value function parameters $\varphi$
\For{update epoch number $= 0, 1, \dots, K$}
    \State Compute cumulative reward list $R_{N_u \times 1}$ using reward list $\mathcal{R}_{N_u \times 1}$, and evaluate expected returns list $V_{N_u \times 1}$ via observation list $\mathcal{O}_{N_u \times 4}$ and expected returns function $V_{\varphi}$
    \State Compute the advantage function list: $A_{N_u \times 1} = R_{N_u \times 1} - V_{N_u \times 1}$
    \State Evaluate $\pi_\vartheta$ using observation list $\mathcal{O}_{N_u \times 4}$ and action list $\mathcal{A}_{N_u \times 2}$, storing the probability in $\pi_{\vartheta}(\mathcal{A}_k|\mathcal{O}_k)_{N_u \times 1}$
    \State Compute probability ratio: $r(\vartheta) = \frac{\pi_{\vartheta}(\mathcal{A}_k|\mathcal{O}_k)_{N_u \times 1}}{\pi_{\vartheta_{old}}(\mathcal{A}_k|\mathcal{O}_k)_{N_u \times 1}}$
    \State Calculate clipped surrogate loss function $L^{\text{CLIP}}(\vartheta)$, returns loss function $L^{\text{ER}}(\varphi)$, and entropy loss $L^S(\vartheta)$
    \State Compute total loss $L(\vartheta, \varphi) = -L^{\text{CLIP}}(\vartheta) + L^{\text{ER}}(\varphi) - L^S(\vartheta)$
    \State Update parameters $(\vartheta, \varphi)$ using Adam optimizer to minimize total loss
\EndFor
\end{algorithmic}
\end{algorithm}

\section{The impact of training and reward parameters on the swimmer's performance}

\begin{figure*}[t]
\centering
 	\includegraphics[scale=0.8]{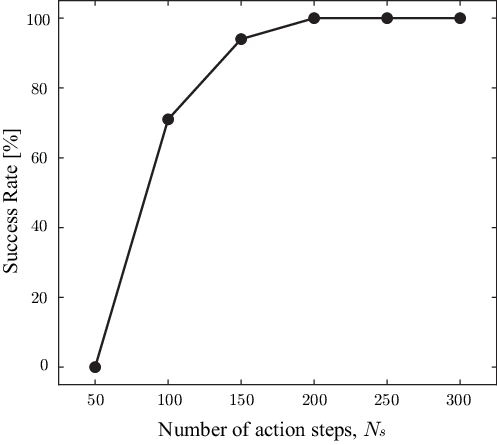}
 	\caption{\footnotesize Variation of success rate as a function of $N_s$ based on 100 trials. The models are trained with the VFS reward function. Here, $N_E=100,000$.}
 	\label{fig:Ns}
\end{figure*}

Fig.~\ref{fig:Ns} illustrates the effect of the number of action steps per episode ($N_s$) on the success rate of navigation, based on models trained with the VFS reward function. A trial is considered successful if the average orientation during the translational stage is below the threshold of $|\theta_s|<2.5^{\text{o}}$. While increasing $N_s$ leads to longer training times, a sufficiently large value is necessary to achieve a high success rate. We set $N_s=200$, which yields a 100\% success rate while keeping training time minimal.

\begin{figure*}[t]
\centering
 	\includegraphics[scale=0.8]{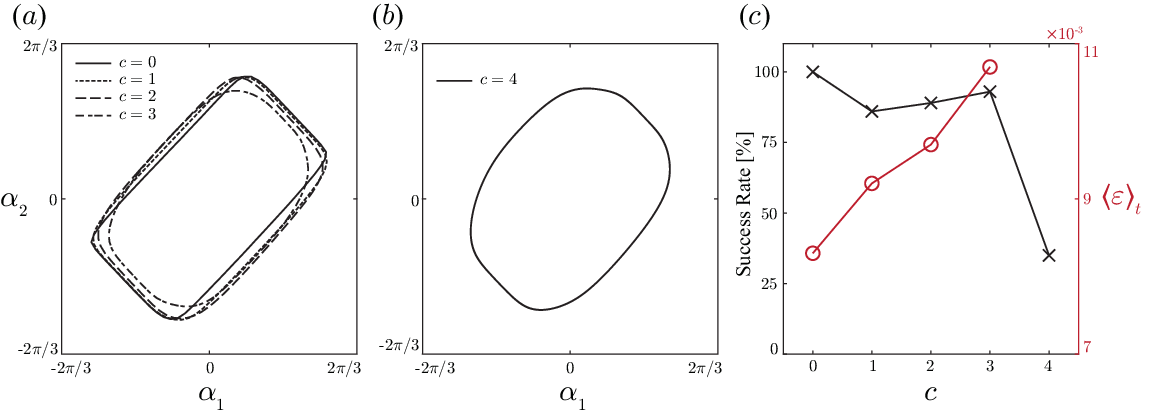}
 	\caption{\footnotesize The impact of the penalty weight on the swimmer's performance.  $(a)$ Stroke patterns for \( c = 0, 1, 2, 3 \).  $(b)$ Stroke patterns for \( c = 4 \).  $(c)$ Left: Success rate in a navigation test for various \( c \) values (black line with cross markers) based on 100 trials. Right: Swimming efficiency for different \( c \) values (red line with circle markers). In all cases, \( b = 6 \) is maintained.}
 	\label{fig:ratio_c/b}
\end{figure*}

In Fig.~\ref{fig:ratio_c/b}, we demonstrate the impact of the penalty weight \(c\) on the swimmer’s performance. In Fig.~\ref{fig:ratio_c/b}($a$), we compare the stroke patterns obtained for different values of \(c\). When \(c = 1\) or \(c = 2\), the stroke patterns show little deviation from those produced by the Velocity-Focused Strategy (i.e., \(c = 0\)). However, at \(c = 3\) the stroke patterns become noticeably more rounded. In Fig.~\ref{fig:ratio_c/b}($c$), we quantitatively present the swimming efficiency for the various strategies, showing that efficiency increases as the penalty weight \(c\) increases (indicated by the red line with circle markers).

However, when \(c = 4\), the stroke patterns lose their symmetry (as shown in Fig.~\ref{fig:ratio_c/b}$b$); this is likely due to the swimmer prioritizing energy conservation over advancing toward the target, which decreases navigation accuracy. In Fig.~\ref{fig:ratio_c/b}($c$), we conducted 100 trials for each strategy, and the result indicates that for \(c \leq 3\), the success rate remains high (around 90\%), whereas for \(c = 4\) the disruption in stroke symmetry leads to a marked decrease in performance (in black line with cross markers).